\documentclass[12pt]{article}
\input{bayesuvius.sty}

\begin{document}
\title{Causal DAG extraction from \\
a library of books or videos/movies}
\date{ \today}
\author{Robert R. Tucci\\
        tucci@ar-tiste.com}
\maketitle
\vskip2cm
\section*{Abstract}
Determining a causal DAG (directed acyclic graph) for a problem under
consideration,
is a major roadblock when doing Judea Pearl's
Causal Inference (CI) in Statistics. The same problem arises when doing
CI in Artificial Intelligence (AI) and Machine Learning (ML).
As with many problems in Science,
we think Nature has found an effective solution to 
 this problem.
We argue that human and animal
brains 
contain an explicit 
engine for doing CI,
and that such an engine
uses as input an atlas (i.e., collection)
of causal DAGs.
We propose a simple 
algorithm for constructing 
such an atlas 
from a library of books or videos/movies.
We illustrate our method by applying it
to a database of randomly generated 
Tic-Tac-Toe games. The software used to 
generate this Tic-Tac-Toe example is open source and
available at GitHub.
\newpage

\chapquote{``If humans were so good at causal inference, religion would not exist."}
{Yann LeCun}{Ref.\cite{yann-text}}

\chapquote{``How much of human knowledge is captured in all the text ever written? To which my answer is: not much.
"}{Yann LeCun}{Ref.\cite{yann-religion}}

\section{Introduction}
Automated Causal  DAG Extraction From Text (DEFT)
seems like a daunting task today,
at a time when little work has been 
done to achieve it. 
Some previous work to achieve DEFT can be 
found in Ref.\cite{2022opberg, waltz}. Once  DEFT is achieved,
the extracted DAGs can be 
collected together as an atlas (i.e., collection) of DAGs. This DAG atlas can be used 
as input to a
Causal Inference (CI) engine
within AI/ML software.
The job of such a CI engine 
will be to
distinguish between
correlation and causation,
and to
perform all 3 rungs (prediction,
do operators, counterfactuals)
of Judea Pearl's ladder of CI. (See Ref.
\cite{book-of-why}).

Contrary to what naysayers
like Yann LeCun predict,
we believe an 
explicit CI engine 
will some day be 
an essential part of all AI
software. 

We also believe that,
contrary to what Yann LeCun claims
in the above quote,
there is ample evidence
that a very efficient CI engine already 
exists in all human and animal 
brains. Maybe not 
all human brains 
excel at CI,
but some clearly have (e.g., Einstein's
and Feynman's). 
And if human/animal brains possess an explicit
CI engine, it's not 
too far fetched to assume
that they also possess a DAG atlas
to provide input to that CI engine.
There even seems to be evidence that 
humans/animals
are not born as a Tabula Rasa (blank slate),
but possess at birth a primitive DAG atlas.
For example, one can find on YouTube
videos showing how human babies
are born with an aversion to 
sitting on grass.

Some people, like
Yann LeCun,
 believe that
human/animal brains, rather than 
possessing an explicit CI 
engine, perform
CI through spontaneous generation, or
random chance, or
Divine Intervention.
We vehemently disagree.
An AI without
an explicit CI engine
would be DIM (Divine
Intervention Mechanism)
witted, and would
not act like the
human or animal mind.
If at the dawn of life,
human and animal brains
had lacked an explicit CI engine,
Nature,
through evolution, would have invented it early on. The 
survival
advantage of 
such a CI engine
is enormous.

Finally,
contrary to 
what LeCun
claims in the above quote,
the  knowledge 
captured in all the text
ever written,
is enormous. 
One of 
the goals of this
paper is to 
improve AI by
mining the vast
trove of
human knowledge 
(especially, causal knowledge)
contained in text (and other 
sources like videos).
Once that vast 
trove of human knowledge
can be mined in an automated 
fashion, this will
open the floodgates 
for implementing CI in AI/ML and Statistics.
As Judea Pearl has said,
CI will elevate AI 
from Deep Learning
to Deep Understanding.

We claim that the DEFT algorithm 
proposed in this paper
extracts {\it causal} DAGs. This
begs the question, what
is causality? This is 
how I define causality in my book
``Bayesuvius" (Ref.\cite{bayesuvius}) 
 
\begin{quote}
Causality is a time-induced ordering between two events, the transmission of information (and its accompanying energy) from the earlier of the two events to the later one, and the physical response of the later event to the reception of that information.
\end{quote}
I expand more on this definition in Bayesuvius, 
in its Appendix E
entitled: ``Bayesian Networks, Causality
and the Passage of Time".

\section{Proposal for
a simple DEFT algorithm}

In this section,
we will present 
our proposal
for a simple DEFT
algorithm.
We will
use the acronym 
DEFT to mean extraction
of DAGs
from various primary sources.
By {\bf primary sources},
we will
mean, not
just text (e.g., libraries of
books),
but also libraries of videos
and other sources or
a mixture of these.

The first step in
our algorithm
is to covert 
the primary sources
into a library
of one or more comic books (CBs).
By a {\bf comic book (CB)},
we mean
a book that
consists of 
a finite number
of chronologically 
ordered {\bf frames}.
Different CBs 
in the 
library may
contain a different
number of frames.
Each frame in a CB
must be annotated
with an {\bf event descriptor}
for each of
the most important {\bf events}
occurring in the frame.
An example
of an event descriptor
might be ``Boy kicks soccer ball".
It's just a very simple sentence
with at least a subject and a verb,
and, occasionally, 
more info than a subject-verb.

Reduction of primary
sources to a library of CBs
might be
done completely
by machine learning
software (fully-automated)
or completely by humans
or both.
There already
exist various 
image-to-text and text-to-image
conversion
software apps.
If the
primary sources
are a library of
movies,
perhaps the
image-to-text
conversion
apps can be
modified to
select 
a finite number
of chronological ordered frames and 
to annotate them
with event descriptors.
If the primary sources are 
a library of books,
perhaps
the text-to-image
apps  can be modified 
to produce the CBs.
Hence, it's possible that 
someday a fully automated
DEFT algorithm will 
be achieved.

The
reason
why we  do primary 
sources to CBs
conversion
as the 
first step
of our DEFT
algorithm
is that
a primary 
source like a
text often
has
the chronological
order of events
implicit.
To mine
the causal
info of a primary
source, we first need to 
make its chronological
order of events 
very clear and explicit.
There is much more
to causality than chronological
ordering,
but chronological 
ordering is an intrinsic
component of it.

The second step
in our DEFT
algorithm 
is to convert
the 
library of CBs
produced in the first step
into a collection of DAGs
({\bf DAG atlas}). We shall
describe this second step
using a simple fictional
example.

For our fictional example,
we shall end up
considering a {\bf number of events}
$N_{e}=11$. The event
descriptors are as follows.
Note that sometimes,
to save space,
we will
abbreviate the 
event descriptor 
by a 4 letter symbol.

\newcommand{\RoCr}[0]{\text{Rooster crows}}
\newcommand{\SuRi}[0]{\text{Sun rises}}
\newcommand{\BuSt}[0]{\text{Bus starts daily rounds}}
\newcommand{\BoWa}[0]{\text{Boy wakes up}}
\newcommand{\BoEa}[0]{\text{Boy eats breakfast}}
\newcommand{\BoRu}[0]{\text{Boy runs to meet bus}}
\newcommand{\BuAr}[0]{\text{Bus arrives at boy's house}}
\newcommand{\DoBa}[0]{\text{Dog barks}}
\newcommand{\BoBo}[0]{\text{Boy boards bus}}
\newcommand{\RaSt}[0]{\text{Rain starts}}
\newcommand{\RaEn}[0]{\text{Rain ends}}

\begin{enumerate}
\item{\it
SuRi:} \SuRi
\item{\it
RoCr:} \RoCr
\item{\it
BuSt:} \BuSt
\item{\it
BoWa:} \BoWa
\item{\it
BoEa:} \BoEa
\item{\it
BoRu:} \BoRu
\item{\it
BuAr:} \BuAr
\item{\it
DoBa:} \DoBa
\item{\it
BoBo:} \BoBo
\item{\it
RaSt:} \RaSt
\item{\it
RaEn:} \RaEn
\end{enumerate}

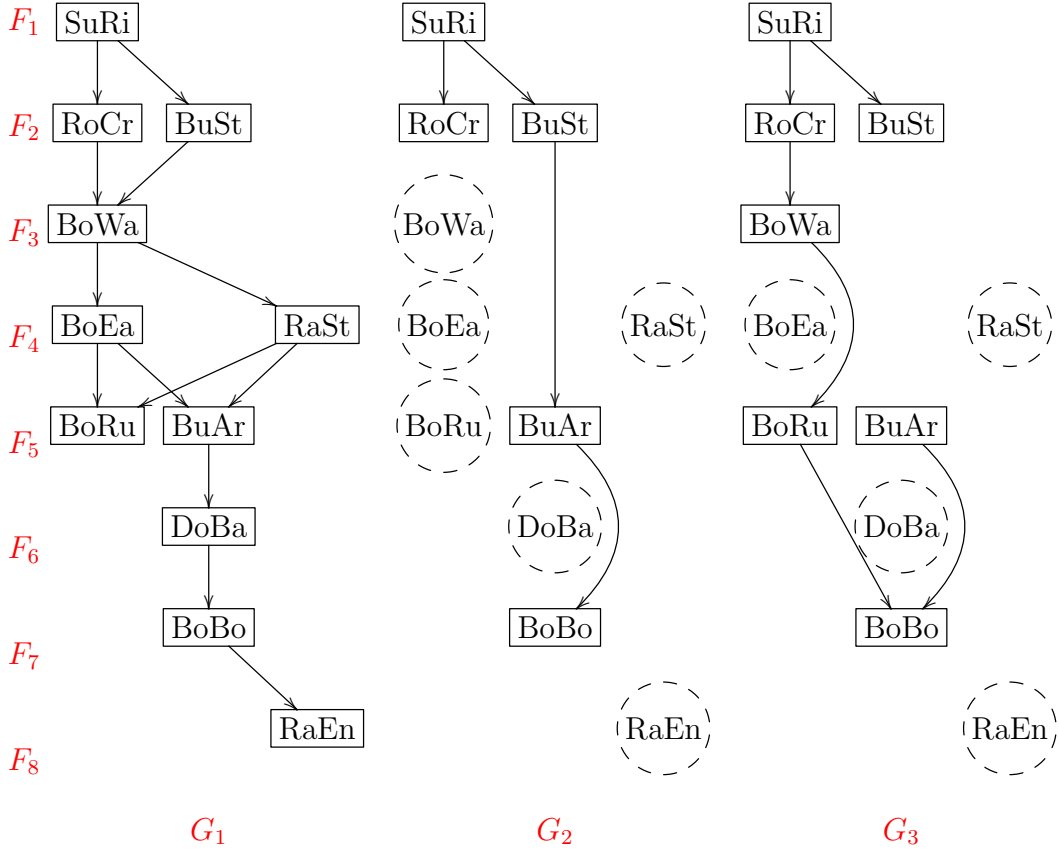
\begin{figure}[h!]
$$
\xymatrix@C=.5pc{
{\color{red}F_1}
\\
{\color{red}F_2}
\\
{\color{red}F_3}
\\
{\color{red}F_4}
\\
{\color{red}F_5}
\\
{\color{red}F_6}
\\
{\color{red}F_7}
\\
{\color{red}F_8}
}
\xymatrix@C=.5pc{
*+[F]{\rm SuRi}\ar[d]\ar[dr]
\\
*+[F]{\rm RoCr}\ar[d] & *+[F]{\rm BuSt}\ar[dl]
\\
*+[F]{\rm BoWa}\ar[d]\ar[drr]
\\
*+[F]{\rm BoEa}\ar[dr]\ar[d]
&&*+[F]{\rm RaSt}\ar[dl]\ar[dll]
\\
*+[F]{\rm BoRu}
&*+[F]{\rm BuAr}\ar[d]
\\
&*+[F]{\rm DoBa}\ar[d]
\\
&*+[F]{\rm BoBo}\ar[dr]
\\
&&*+[F]{\rm RaEn}
\\
&\color{red}G_1
}
\quad 
\xymatrix@C=.5pc{
*+[F]{\rm SuRi}\ar[d]\ar[dr]
\\
*+[F]{\rm RoCr} 
& *+[F]{\rm BuSt}\ar[ddd]
\\
*+[F-o]{\rm BoWa}
\\
*+[F-o]{\rm BoEa}
&&*+[F-o]{\rm RaSt}
\\
*+[F-o]{\rm BoRu}
&*+[F]{\rm BuAr}\ar@/^2pc/[dd]
\\
&*+[F-o]{\rm DoBa}
\\
&*+[F]{\rm BoBo}
\\
&&*+[F-o]{\rm RaEn}
\\
&\color{red}G_2
}
\quad 
\xymatrix@C=.5pc{
*+[F]{\rm SuRi}\ar[d]\ar[dr]
\\
*+[F]{\rm RoCr}\ar[d] 
& *+[F]{\rm BuSt}
\\
*+[F]{\rm BoWa}\ar@/^2pc/[dd]
\\
*+[F-o]{\rm BoEa}
&&*+[F-o]{\rm RaSt}
\\
*+[F]{\rm BoRu}\ar[ddr]
&*+[F]{\rm BuAr}\ar@/^2pc/[dd]
\\
&*+[F-o]{\rm DoBa}
\\
&*+[F]{\rm BoBo}
\\
&&*+[F-o]{\rm RaEn}
\\
&\color{red}G_3
}
$$
\caption{
Let 
$\gamma=1,2, 3, \ldots, N_{cb}$
label the CBs in
our current cbLib$^*$.
For the $\gamma$'th 
CB, we construct
a DAG  $G_\gamma$.
  Here we show  the 
  DAGs for the first
  3 CBs
  of our current cbLib$^*$.
 Each row $F_i$ for 
 $i=1,2, \dots, N_f$
 are the frames of the CBs.
 Nodes enclosed by 
 a rectangle
 represent events that
 happened
 in a CB.
Nodes enclosed by a dashed circle
are
placeholders for
events that didn't happen
in a CB.}
\label{fig-dags123}
\end{figure}

Suppose two CBs $c_1$ and $c_2$
have event sets $E(c_1)$ and $E(c_2)$,
respectively. 
We say {\bf $c_1$ is time compatible (tc) smaller than $c_2$ ($c_1 < c_2$)}
if (1) $E(c_1)$ is a proper subset of $E(c_2)$
and (2) the events in $c_1$
are in the same 
chronological order as the corresponding
events in $c_2$.

We define a {\bf time compatible CB library (cbLib$^*$)}
to be a library (i.e., set) of CBs wherein
there is one CB of maximum
length, and all other CBs in the library are tc-smaller
than the maximum length one.

A {\bf CB $c$
is time compatible
with a cbLib$^*$ $C$}
 if 
 $c<c_{max}$, where $c_{max}$ is the
 maximum length CB of $C$.\footnote{Another possible definition for this
 is that $c<c_{max}$ or $c>c_{max}$.}

Given a large set of CBs,
we can generate a collection $\calc$
of cbLib$^*$s as follows.
If $\calc$ is empty, find a CB $c$,
then create a cbLib$^*$ 
with $c$ as its only CB,
and add that cbLib$^*$ to
$\calc$.
If $\calc$ is not empty,
and we want to add a new CB called $c$ to
$\calc$,
add $c$ to every cbLib$^*$ in
 $\calc$ with which it is time compatible.
 If we can't find a cbLib$^*$ in $\calc$
 with which $c$ is time compatible, then
 create a new cbLib$^*$ with $c$
 as its only member
 and add it to $\calc$.
 
 Next we shall explain 
 how to process each cbLib$^*$ 
 individually.

Let 
$\gamma=1,2, 3, \ldots, N_{cb}$
label the CBs in
our current cbLib$^*$.
For the $\gamma$'th 
CB, we construct
a DAG  $G_\gamma$.
Fig.\ref{fig-dags123}
shows,
for our
fictional example, the DAGs for
the first 3 CBs of our
current cbLib$^*$.
The DAGs in 
Fig.\ref{fig-dags123}
were constructed
very easily by
placing the event
descriptors 
in chronological order
with time
pointing down the page,
and adding arrows 
entering each event $e$
from all events $e'$
in the frame immediately before
$e$'s frame.
We will refer to 
this as $T_{mem}=1$ {\bf memory time}.
We could have
just as well 
added arrows entering each event $e$
from all events $e'$
in the two frames 
immediately before 
$e$'s frame.
This would be the case $T_{mem}=2$.
In general,
one could use any $T_{mem}$,
including  $T_{mem}$
as big as the total number 
of frames,
but the higher $T_{mem}$
is, the more dense the DAG will be.
I think $T_{mem}$ 
equal to 1 or 2 will
provide a good enough causal fit
most of the time.

\begin{figure}[h!]
$$
\xymatrix{
*+[F]{\rm \SuRi}\ar[d]^{80}\ar[dr]^{98}
\\
*+[F]{\rm\RoCr}\ar[d]^{60} & *+[F]{\rm\BuSt}\ar[ddd]^{55}
\\
*+[F]{\rm\BoWa}\ar[d]_{52}\ar@/^6pc/[dd]^{50}
\\
*+[F]{\rm\BoEa}
&&*+[F]{\rm\RaSt}\ar[dddd]^{54}
\\
*+[F]{\rm\BoRu}\ar[ddr]^{67}
&*+[F]{\rm\BuAr}\ar[d]^{60}\ar@/^5pc/[dd]^{50}
\\
&*+[F]{\rm\DoBa}
\\
&*+[F]{\rm\BoBo}
\\
&&*+[F]{\rm\RaEn}
}
$$
\caption{
High frequency arrows DAG, $G_{hfa}$,
 with $N_{art} = 50$.
Arrow repetition number
given besides each arrow.
Fig.\ref{fig-dags123}
shows DAGs 
$G_\gamma$ for $\gamma=1,2,3$.
}
\label{fig-hf-dag}
\end{figure}
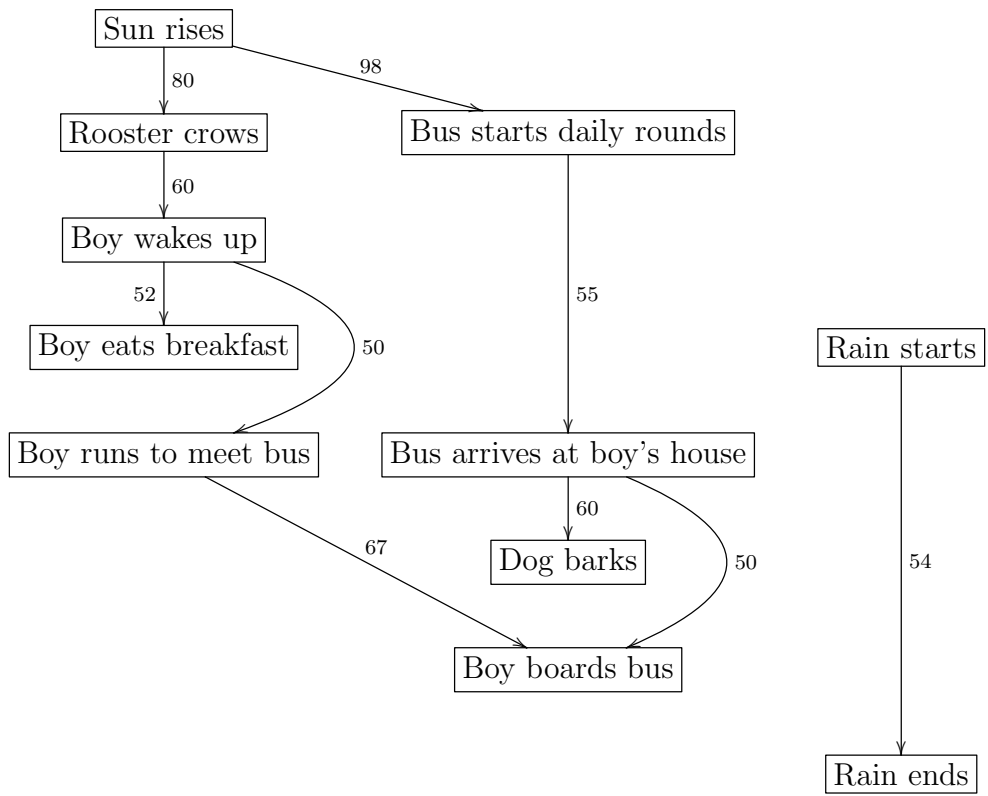

At this point, we are ready 
to calculate $G_{hfa}$,
the {\bf high frequency arrows (hfa)} DAG.
This DAG is created
by counting the 
number of times each arrow
appears in the 
just created DAGs $G_\gamma$
for all $\gamma$, 
and keeping 
only arrows that are repeated
at least as much as
the {\bf arrow repetition threshold} $N_{art}$. 
Fig.\ref{fig-hf-dag}
shows a possible $G_{hfa}$
for our fictional 
example, assuming $N_{art}=50$.
The intuition behind
$G_{hfa}$ is that
arrows that have low repetition numbers do so because Nature is telling us that they are not essential for the causal mechanism being expressed by the whole DAG.

Note that one can upgrade $G_{hfa}$
from a causal DAG to
a causal Bayesian Network (bnet)
by calculating  an 
empirical estimate 
for the
 TPM (Transition Probability Matrix)
 of each event node $\rve\in \bool$
 of $G_{hfa}$.
 This estimate can be calculated using the following 
 formula.
 
 \beq
 P(e|pa(\rve)= \pi )=
 \frac{\sum_{\gamma=1}^{N_{cb}}
 \indi(
 \rve_\gamma=e,[ pa(\rve)]_\gamma = \pi)
 }{
 \sum_{\gamma=1}^{N_{cb}}
 \indi(
 [pa(\rve)]_\gamma = \pi)
 }
 \eeq
 Here $e\in \bool$ and 
 $\pi\in \bool^n$,
 where $n$
 is the number of parents of node $\rve$
 of $G_{hfa}$.
 Also, node $\rve$ in $G_{hfa}$
 and node $\rve_\gamma$ in $G_\gamma$
 refer to the same event.\footnote{
 We are using the 
 convention
 used in Bayesuvius (Ref.\cite{bayesuvius})
 of indicating random  variables
 by underlining them,
 and using $pa(\rvx)$
 to denote the parent nodes
 of node $\rvx$.}
 
Note that $G_{hfa}$
might consist of more than
one connected DAG.
Furthermore, note 
that we
have shown
how to generate 
a $G_{hfa}$ for each cbLib$^*$ 
in
a collection $\calc$
of cbLib$^*$s.
Hence, this algorithm
generates a collection 
of $G_{hfa}$ DAGs.
This collection
of DAGs is what we call
the {\bf DAG atlas}.

Note that our DEFT algorithm is very simple and only requires counting, 
and keeping good chronologically
ordered records. We believe that
the human brain excels at those tasks. Our DEFT algorithm avoids performing precise numerical calculations like calculating $\pi$ to 15 decimal places because we believe most human brains can't perform such tasks.

\section{How to use the DAG Atlas}
Once a DAG atlas is calculated,
and its DAGs are upgraded to bnets, 
it can be used to 
perform predictions (1st rung),
do operators (2nd rung),
and to calculate CATE and 
ATE for counterfactuals (3rd rung).
The results of these calculations
can be used by the AI to make decisions.

In the particular version 
of the 3 rungs that is advocated
in Bayesuvius \footnote{See the chapter
entitled ``Counterfactual Reasoning" in
Bayesuvius, Ref.
\cite{bayesuvius}},
the 2nd rung removes arrows from a bnet
using a do-operator,
and the 3rd rung adds new nodes 
and arrows to a bnet
using an imagine operator.
My point is that the DAG atlas
can be modified by applying
the rules of CI.

Note that 
hidden nodes
have not been mentioned
so far. Recall
that hidden nodes are
nodes in a DAG
that one would like
to include 
because we think
they are important, but
which haven't been
measured, 
either because
it is impossible to 
measure them, 
or we simply haven't gotten
around to measuring them.
Hidden nodes could certainly
be added to the DAGs of a DAG 
altas,
either in an automated fashion,
or by hand.

Note that if a CB repeats 
a task over and over again,
the DAG $G_{hfa}$
might turn out
to be a dynamical Bayesian network.\footnote{
Dynamical Bayesian networks are discussed in
Bayesuvius, Ref.\cite{bayesuvius}}

\appendix
\section{Appendix: Tic-Tac-Toe example}
See Ref.\cite{tic-tac-toe} for Python code
that implements,
for Tic-Tac-Toe, the DEFT algorithm 
proposed in this paper.

A CB is a list of 
chronologically ordered frames. For the Tic-Tac-Toe example considered in 
this code, a CB is one Tic-Tac-Toe game, and a frame is just one move. For example, 

\beq\rm
['X1',\; 'O2',\; 'X8',\; 'O0',\; 'X6',\;
 'O4',\; 'X7']
\label{eq-ttt-game} 
\eeq
represents a 
CB/game and 'X1' is a frame/move. The X and O refer to the player. X always 
plays first. The numbers refer to the positions on the Tic-Tac-Toe grid, 
labelled as follows: 
\beq
\setlength\arrayrulewidth{2pt}\begin{tabular}{c|c|c}0&1&2\\\hline 3&4&5\\\hline 6&7&8\end{tabular}    
    \eeq

In the more general case considered in this paper, a frame, instead of being 
a single string, can be a list of strings called event descriptors, 
which represent simultaneous events in the frame. 

The time each X or O is played will be indicated by the time enclosed in
    parenthesis. For example, X3(2) is an X played at time 2 and position 3.
    The 3, which indicates the position on the Tic-Tac-Toe grid,
    is redundant and omitted if X3(2) is rendered inside a Tic-Tac-Toe grid.
    For example, here is the game Eq.(\ref{eq-ttt-game})
    represented on a grid.
    
    \beq
\setlength\arrayrulewidth{2pt}\begin{tabular}{c|c|c}O(3)&X(0)&O(1)\\\hline &O(5)&\\\hline X(4)&X(6)&X(2)\end{tabular}    
    \eeq
    
Our code generates
2000  CBs at random, 
and calculates a collection $\calc$ 
of cbLib$^*$s from 
those. For each cbLib$^*$ $C\in \calc$, 
it can draw a DAG 
for each CB $c\in C$,
for a given memory time
$T_{mem}$ .
It can then calculate 
the high frequency arrows DAG
$G_{hfa}$ of $C$, for
a given arrow repetition threshold $N_{art}$.
It can also upgrade $G_{hfa}$
to a bnet. 

\begin{figure}[h!]
\centering
\includegraphics[width=5in]
{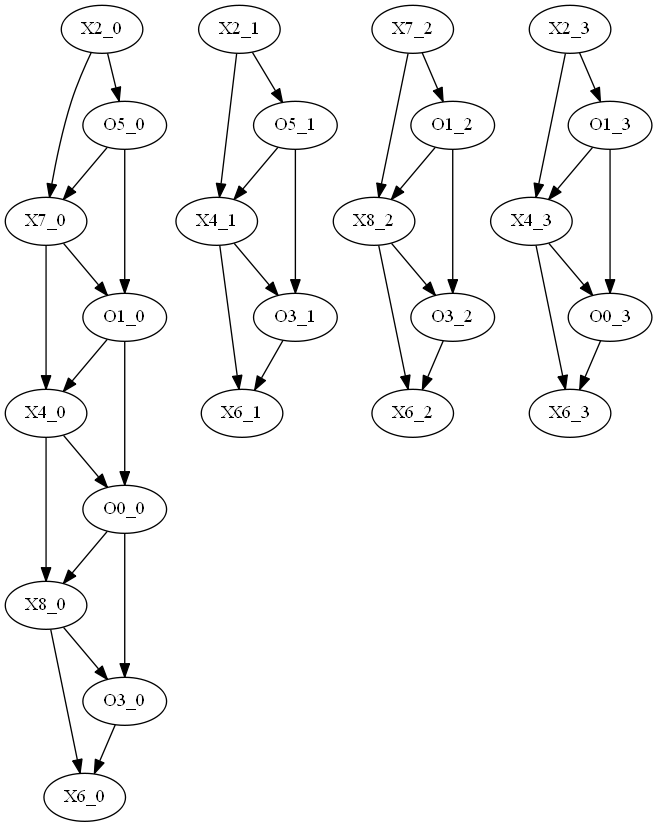}
\caption{DAGs for the cbLib$^*$ 
of Eq.(\ref{eq-cbLibX-typical}),
with memory time $T_{mem}=2$. }
\label{fig-dags-cbLibX}
\end{figure}

\begin{figure}[h!]
\centering
\includegraphics[width=2.5in]
{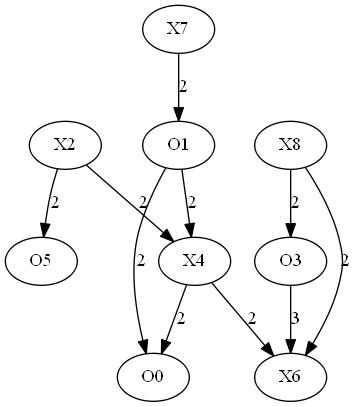}
\caption{High frequency arrows graph 
$G_{hfa}$ with 
arrow repetition threshold $N_{art}=2$}
\label{fig-G-hfa}
\end{figure}

As an example, consider the following
cbLib$^*$:

\beq\footnotesize
\begin{tabular}{cccc}\setlength\arrayrulewidth{2pt}\begin{tabular}{c|c|c}O(5)&O(3)&X(0)\\\hline O(7)&X(4)&O(1)\\\hline X(8)&X(2)&X(6)\end{tabular}&\setlength\arrayrulewidth{2pt}\begin{tabular}{c|c|c}&&X(0)\\\hline O(3)&X(2)&O(1)\\\hline X(4)&&\end{tabular}&\setlength\arrayrulewidth{2pt}\begin{tabular}{c|c|c}&O(1)&\\\hline O(3)&&\\\hline X(4)&X(0)&X(2)\end{tabular}&\setlength\arrayrulewidth{2pt}\begin{tabular}{c|c|c}O(3)&O(1)&X(0)\\\hline &X(2)&\\\hline X(4)&&\end{tabular}\end{tabular}
\label{eq-cbLibX-typical}
\eeq
The DAGs for this cbLib$^*$
are given in Fig.\ref{fig-dags-cbLibX}.
Its high frequency arrows DAG $G_{hfa}$
is given by Fig.\ref{fig-G-hfa}.
And the TPMs of $G_{hfa}$
are calculated by our software
to be as follows.
The TPM of a node $\rvx$ is expressed as
a Python 
dictionary 
mapping the state
of the parents of $\rvx$ to
a pair of floats
that sum to one and
give the probability that node $\rvx$ is 
either in state 0 (absent)
or 1 (present).
Some TPM entries
are undefined,
a problem which 
might be ameliorated with
more data.

\begin{enumerate}
\item node: X2
, parents: []
$$TPM = \left\{ \begin{array}{ll}():& [inf, inf]\end{array}\right\}$$

\item node: O5
, parents: ['X2']
$$TPM = \left\{ \begin{array}{ll}(0,):& [1., 0.],\\  (1,):& [0.33333333, 0.66666667]\end{array}\right\}$$

\item node: X7
, parents: []
$$TPM = \left\{ \begin{array}{ll}():& [inf, inf]\end{array}\right\}$$

\item node: O1
, parents: ['X7']
$$TPM = \left\{ \begin{array}{ll}(0,):& [0.5, 0.5],\\  (1,):& [0., 1.]\end{array}\right\}$$

\item node: X4
, parents: ['O1', 'X2']
$$TPM = \left\{ \begin{array}{ll}(0, 0):& [nan, nan],\\ 
 (0, 1):& [0., 1.],\\ 
 (1, 0):& [1., 0.],\\ 
 (1, 1):& [0., 1.]\end{array}\right\}$$

\item node: O0
, parents: ['X4', 'O1']
$$TPM = \left\{ \begin{array}{ll}(0, 0):& [nan, nan],\\ 
 (0, 1):& [1., 0.],\\ 
 (1, 0):& [1., 0.],\\ 
 (1, 1):& [0., 1.]\end{array}\right\}$$

\item node: X8
, parents: []
$$TPM = \left\{ \begin{array}{ll}():& [inf, inf]\end{array}\right\}$$

\item node: O3
, parents: ['X8']
$$TPM = \left\{ \begin{array}{ll}(0,):& [0.5, 0.5],\\  (1,):& [0., 1.]\end{array}\right\}$$

\item node: X6
, parents: ['O3', 'X8', 'X4']
$$TPM = \left\{ \begin{array}{ll}(0, 0, 0):& [nan, nan],\\ 
 (0, 0, 1):& [0., 1.],\\ 
 (0, 1, 0):& [nan, nan],\\ 
 (0, 1, 1):& [nan, nan],\\ 
 (1, 0, 0):& [nan, nan],\\ 
 (1, 0, 1):& [0., 1.],\\ 
 (1, 1, 0):& [0., 1.],\\ 
 (1, 1, 1):& [0., 1.]\end{array}\right\}$$
\end{enumerate}

\bibliographystyle{plain}
\bibliography{references}
\end{document}